\documentclass{article}

\usepackage{booktabs}
\usepackage{graphicx}
\usepackage{algorithm}
\usepackage{algpseudocode}
\usepackage{subcaption}
\usepackage{wrapfig}
\usepackage{enumitem}
\usepackage{hyperref}
\usepackage{fontawesome5}
\usepackage{amsmath}
\usepackage{amsfonts}
\usepackage{amssymb}
\usepackage{multirow}
\usepackage[preprint]{corl_2026}

\definecolor{darkgreen}{rgb}{0.0, 0.5, 0.0}

\newcommand{\modelname}{{\textsc{Dex-X}}}

\title{Learning Visual-Tactile Dexterous Manipulation From Human Videos with Simulated Interaction}

\author{%
\makebox[\textwidth][c]{%
\begin{minipage}{1.14\textwidth}
\centering
\normalfont\normalsize
\vspace{2.5mm}
\textbf{%
Ruoqu Chen$^{1,2}$,
Feixiang Ruan$^{1,3,4,*}$,
Liu Cao$^{1,2}$,
Zihao Wang$^{1}$,
Botian Xu$^{1,*}$,
Shiqin Tong$^{1,*}$%
}\\[1mm]
\textbf{%
Jiajun Liu$^{1,2,5,*}$,
Mingzhi Pei$^{1}$,
Chenyu Zhang$^{1,2}$,
Wanli Xing$^{3}$,
Kaifeng Zhang$^{3}$,
Mengdi Xu$^{1,\dagger}$%
}\\[2mm]
$^{1}$IIIS, Tsinghua University
\qquad
$^{2}$Shanghai Qizhi Institute
\qquad
$^{3}$Sharpa
\\[0.5mm]
$^{4}$Tongji University
\qquad
$^{5}$Renmin University
\\[1mm]
$^{*}$Work done during internship at Tsinghua University.
\qquad
$^{\dagger}$Corresponding author.
\\[0.7mm]
\href{https://dexx-code.github.io/dexx-code/}
{\faGlobe\ \texttt{dexx-code.github.io/dexx-code}}
\vspace{-5.5mm}
\end{minipage}%
}%
}

\begin{document}
\maketitle


\begin{abstract}
    Human videos are an abundant source of dexterous manipulation behaviors, but they lack tactile information that is crucial for contact-rich interaction. This raises a fundamental question: 
    can robots learn deployable visual-tactile dexterous manipulation policies from human video demonstrations without robot-side data collection?
    We present \modelname, a framework for learning visual-tactile dexterous manipulation from human videos through simulation. Our key insight is that simulation can serve as a tactile completion engine. Given monocular human demonstrations, \modelname{} reconstructs hand-object interactions in simulation, where physically grounded contact dynamics provide tactile supervision unavailable in the original videos. Leveraging this recovered tactile information, we train visual-tactile dexterous manipulation policies and distill them into deployable policies operating on point-cloud observations and tactile sensing.
    We demonstrate zero-shot sim-to-real transfer on a dexterous hand-arm platform across diverse grasping and contact-rich tool-use tasks. 
    The teacher policy achieves 65.9\% average success across six task categories in simulation, while the distilled visual-tactile policy achieves 93\% success on real-world cube picking and 53\% on the challenging table-cleaning task. Zero-shot generalization to unseen object geometries is also observed on object-picking tasks.
    Our results suggest that simulated interaction is a key bridge between human videos and deployable dexterous manipulation policies, providing the missing physical supervision needed for scalable robot skill learning from Internet-scale human video data.
    More visualizations are on the project website: \href{https://dexx-code.github.io/dexx-code/}{https://dexx-code.github.io}.

\end{abstract}

\keywords{Visual-Tactile Dexterous Manipulation, Reinforcement Learning, Learning from Human Demonstrations} 


\begin{figure}[h]
    \centering
    \includegraphics[width=\linewidth]{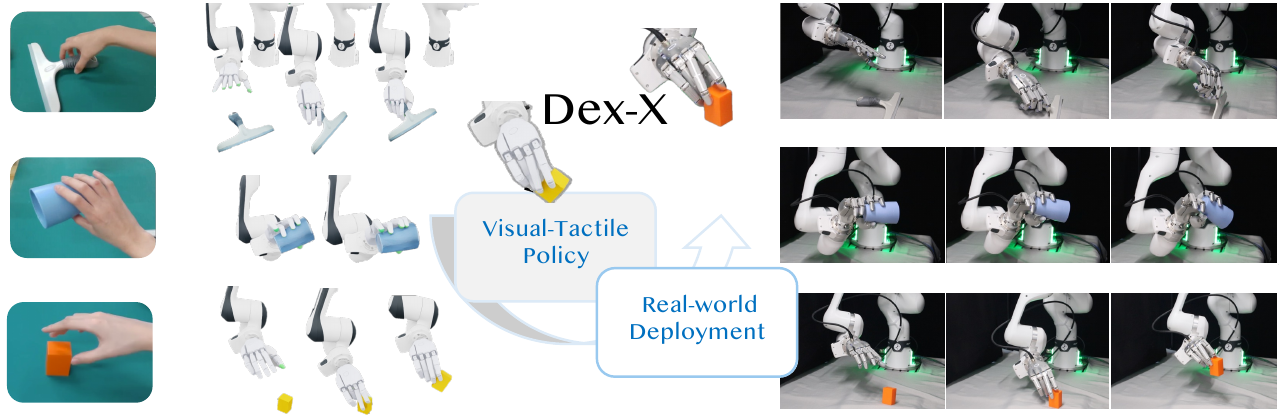}
    \caption{
    \textbf{\modelname{} Overview}. \modelname{} learns deployable visual-tactile dexterous manipulation policies from human video demonstrations. Reconstructed human hand-object references guide interaction learning in simulation, where contact dynamics provide the missing force-level tactile supervision. The resulting policies transfer zero-shot to real-world grasping and tool-use tasks.
    }
    \vspace{-0.2in}
    \label{fig:teaser}
\end{figure}





\section{Introduction}

Dexterous manipulation requires robotic systems to coordinate high-dimensional hand-arm motions under complex contact dynamics while continuously adapting to multimodal sensory feedback. Humans perform such behaviors effortlessly in everyday activities, seamlessly integrating vision and touch to grasp objects, manipulate tools, and perform contact-rich tool-use interactions across diverse environments. Replicating this visual-tactile dexterous manipulation capability on robotic platforms remains a long-standing challenge in robotics~\cite{agarwal2023dexterousfunctionalgrasping,yin2025osmoopensourcetactileglove,yin2026emergentdexteritydiverseresets}.

Recent advances in reinforcement learning and imitation learning have enabled increasingly capable dexterous manipulation systems~\cite{handa2024dextremetransferagileinhand,yuan2024crossembodimentdexterousgraspingreinforcement,wang2024lessonslearningspinpens,mandi2025dexmachinafunctionalretargetingbimanual,li2025maniptransefficientdexterousbimanual}. However, both approaches often require either extensive interaction in carefully engineered simulation environments or large-scale robot demonstrations collected through specialized teleoperation systems, making it difficult to scale dexterous manipulation learning to the diversity of real-world interactions. In particular, collecting tactile demonstrations often requires instrumented gloves or custom sensing hardware, which further increases the cost and complexity of data collection.
Human video demonstrations offer a compelling alternative. Videos are abundant and diverse, and capture a vast range of dexterous manipulation behaviors performed in natural environments, making them arguably one of the most scalable sources of manipulation data. This raises a fundamental question: \emph{can robots learn deployable visual-tactile dexterous manipulation policies from human video demonstrations?}

Answering this question requires overcoming two key challenges. First, human videos provide visual observations but lack direct tactile feedback, particularly contact forces that are critical for maintaining stable grasps and sustained object interactions. Second, transferring dexterous manipulation policies from simulation to real hardware remains challenging due to perception noise, actuation delays, and inaccuracies in contact modeling. Together, these challenges make learning deployable visual-tactile policies from human videos an open problem.

In this work, we introduce \modelname, a framework for learning visual-tactile dexterous manipulation policies from human video demonstrations through simulation.  Our key insight is that simulation can serve as a \emph{tactile completion engine}. While tactile observations are absent in human videos, physically grounded interaction in simulation enables the reconstruction of contact-related signals, producing paired visual-tactile trajectories for policy learning. Leveraging these reconstructed tactile observations, \modelname{} learns contact-rich manipulation behaviors through tactile-aware RL.

To bridge simulation and real-world deployment, we distill privileged state-based policies into deployable policies using point-cloud observations, proprioception, and fingertip tactile feedback. Without real-world fine-tuning, the resulting policies achieve 93\% success on cube picking and 53\% on the challenging table-cleaning task, while also exhibiting zero-shot generalization to unseen object geometries. These results demonstrate the potential of simulated physical interaction to bridge passive human demonstrations and real-world dexterous manipulation.

Our contributions are summarized as follows:
\begin{enumerate}[leftmargin=*, itemsep=2pt, topsep=2pt, label=\textbullet]

\item We present \textbf{\modelname{}}, a framework for learning \textbf{visual-tactile dexterous manipulation policies from human video demonstrations}. Our key insight is to use simulation as a \emph{tactile completion engine}, providing physically grounded contact information that is absent from human videos.

\item We introduce a simulation-based teacher-student learning paradigm that combines human-video-derived motion priors with tactile-aware reinforcement learning, enabling the acquisition of contact-rich manipulation skills that transfer zero-shot to real-world deployment.

\item We demonstrate zero-shot sim-to-real transfer on a 29-DoF dexterous hand-arm platform across diverse grasping and tool-use tasks. To the best of our knowledge, this is among the first systems to perform real-world visual-tactile dexterous tool use learned directly from human video demonstrations without task-specific fine-tuning.

\end{enumerate}

\vspace{-0.05in}
\section{Related Works}
\label{sec:related_works}
\vspace{-0.05in}
\textbf{Dexterous Manipulation from Human Demonstrations.}
Human demonstrations have long been used to improve dexterous policy learning, from demonstration-guided reinforcement learning and human grasp affordances~\cite{rajeswaran2018learningcomplexdexterousmanipulation,wu2022learninggeneralizabledexterousmanipulation} to imitation and policy learning from human videos~\cite{qin2022dexmvimitationlearningdexterous,shaw2022videodexlearningdexterityinternet}. A central challenge is bridging the human-to-robot embodiment gap while preserving physically meaningful interactions. Prior work addresses this challenge through motion retargeting, trajectory optimization, and reinforcement learning over reconstructed hand-object trajectories~\cite{li2025maniptransefficientdexterousbimanual,mandi2025dexmachinafunctionalretargetingbimanual,chen2024objectcentricdexterousmanipulationhuman,liu2025dextrackgeneralizableneuraltracking,chen2025vividexlearningvisionbaseddexterous}. Recent approaches further improve contact-consistent reconstruction and enable
real-world policy learning or executable robot trajectories directly from human videos~\cite{chen2026dexterousmanipulationpoliciesrgb,kim2026dexterous}. Our work complements these directions by using retargeted human-motion references to guide reinforcement learning and simulated contact interactions to provide force-level tactile supervision, enabling closed-loop visual-tactile policies that operate with real tactile feedback at deployment.

\textbf{Sim-to-Real Transfer for Dexterous Manipulation.}
Simulation-based reinforcement learning has enabled scalable training of complex robot behaviors and their transfer to real-world systems~\cite{rudin2022learning,zhou2025learning,wang2026learning}. For dexterous manipulation, sim-to-real approaches have enabled increasingly capable manipulation skills~\cite{andrychowicz2020learning,handa2024dextremetransferagileinhand,yin2025learninginhandtranslationusing,wang2024lessonslearningspinpens}. Yet reliable transfer remains challenging, particularly for behaviors sensitive to mismatches in contact dynamics, tactile sensing, and actuation~\cite{zhao2026closingrealitygapzeroshot,pan2026beyond,hsieh2025learning,lou2026dexninja,wang2026r2sevalrobotevaluationrealtosim}. Beyond in-hand manipulation, existing approaches have explored generalizable grasping and broader manipulation behaviors through large-scale simulation, visuomotor policy distillation, and force-aware sim-to-real learning~\cite{singh2025dextrahrgbvisuomotorpoliciesgrasp,yin2026emergentdexteritydiverseresets,lou2026drexdifferentiablerealtosimtorealengine}. Sim-to-real dexterous manipulation has also been extended to dynamic and contact-rich tool use through object-centric goal tracking~\cite{kedia2026simtoolrealobjectcentricpolicyzeroshot}, demonstration-guided grasp adaptation~\cite{Gupta_2026}, and human-object co-tracking~\cite{li2026humanleveldexterousteleoperation}. Our work complements these directions by using retargeted human-motion references to guide interaction learning in simulation, with simulated contact interactions providing force-level tactile supervision for visual-tactile policies that operate with real tactile feedback at deployment.

\textbf{Visual-Tactile Manipulation.}
Tactile sensing provides critical feedback for manipulation when visual observations alone are insufficient. Prior work has explored tactile representation learning~\cite{wu2025canonicalrepresentationforcebasedpretraining}, visual-tactile and force-aware policy learning~\cite{xue2026tubediffusionpolicyreactive,huang2025tactilevlaunlockingvisionlanguageactionmodels,lin2026peelknifealigningfinegrained}, and spatially grounded tactile representations~\cite{huang2026spatiallyanchoredtactileawareness,yuan2024robotsynesthesiainhandmanipulation}.
Most closely related to our representation, Robot
Synesthesia~\cite{yuan2024robotsynesthesiainhandmanipulation} embeds tactile
information into a unified point-cloud representation for zero-shot sim-to-real
in-hand manipulation. Recent approaches further advance contact-rich
manipulation through reactive visual-tactile control, contact-grounded policy
learning, and tactile sim-to-real transfer~\cite{
xue2026tubediffusionpolicyreactive,
xu2026contactgroundedpolicydexterousvisuotactile,
chen2026ptldsimtorealprivilegedtactile,
zhao2026closingrealitygapzeroshot}.
Our work complements these directions by reconstructing human demonstrations
as physical interactions in simulation and using the resulting force-level
tactile feedback to learn autonomous visual-tactile policies for sustained
tool-object-environment interaction.

\begin{figure}[t]
    \centering
    \includegraphics[width=\linewidth]{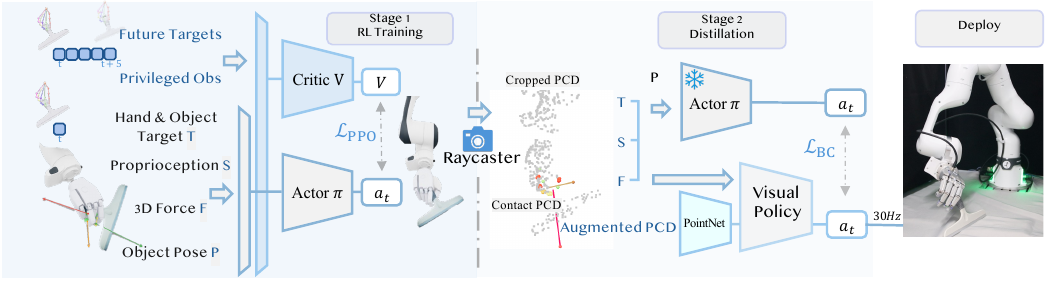}
    \caption{
    \textbf{Training Framework of \modelname.} After transferring human demonstrations into simulation, we train a privileged state-based expert using RL with demonstration references, object states, and tactile contact information. The expert is then distilled into a multi-task visual-tactile policy operating on a unified \emph{contact point cloud} representation, which embeds fingertip tactile feedback into the scene geometry and fuses vision, force, and proprioception for policy learning. The resulting policy transfers zero-shot to diverse real-world dexterous manipulation tasks.
    }
    \label{fig:pipeline}
\end{figure}

\vspace{-0.05in}
\section{\modelname: Learning Visual-Tactile Manipulation from Human Videos}
\label{sec:methods}
\vspace{-0.05in}
In this section, we present \modelname{}, a sim-to-real framework for learning visual-tactile dexterous manipulation from human video demonstrations. The framework consists of three stages. First, human hand-object interactions are reconstructed from video and retargeted into simulation to obtain robot-compatible demonstrations (Sec.\ref{sec:method-retarget}). Second, a privileged-state policy is trained through reinforcement learning guided by these demonstrations, leveraging tactile feedback and contact dynamics unavailable in the original videos (Sec.\ref{subsec:rl}). Finally, the policy is distilled into a deployable visual-tactile controller operating on point clouds, tactile observations, and proprioception, enabling zero-shot transfer to real-world dexterous manipulation tasks (Sec.\ref{subsec:transfer}). Fig.\ref{fig:pipeline} illustrates \modelname{}'s overall training framework.


\subsection{Problem Formulation}
\label{subsec:problem}
We study the problem of learning visual-tactile dexterous manipulation policies from human video demonstrations. Given a dataset of monocular human demonstrations $\mathcal{D} = \{\tau^h_i\}_{i=1}^{N}$
where each trajectory $\tau^h_i =(I_0, I_1, ...)$ consists of a sequence of images capturing a human performing a manipulation task, our goal is to learn a \textit{multi-task visual-tactile policy} that enables a dexterous hand-arm system to reproduce demonstrated behaviors under varying initial conditions.

Formally, we seek a reference-conditioned policy
$\pi(a_t \mid o_t, r_{t+1})$ that maps robot observations $o_t$
and the motion reference $r_{t+1}$ to control actions $a_t$.
The observation
$o_t = \{o_t^{\mathrm{prop}}, o_t^{\mathrm{vis}}, o_t^{\mathrm{tac}}\}$
consists of proprioceptive observations, visual perception, and tactile sensing.

A key challenge is the mismatch between the supervision available in human videos and the sensory feedback required for contact-rich robot execution. Human videos provide visual observations but no direct tactile measurements, whereas successful manipulation often requires contact and force feedback to maintain stable grasps and sustained interaction. We therefore aim to reconstruct the demonstrated behaviors as physical interactions in simulation, obtain the missing tactile supervision from simulated contact dynamics, and learn a unified visual-tactile policy that transfers to real-world execution without further fine-tuning.

\subsection{Motion Prior Data Extraction with Spatial Augmentation}
\label{sec:method-retarget}

\textbf{Hand-Object Motion Reconstruction.}
We first extract human hand-object interaction trajectories from monocular demonstration videos. Given a monocular human demonstration $\tau^h$, we reconstruct the hand-object motion at 30 Hz. Object poses are estimated using FoundationPose~\cite{wen2024foundationposeunified6dpose}, while hand poses are estimated using WiLoR~\cite{potamias2025wilorendtoend3dhand} and further refined with MANO-based temporal consistency and hand-object penetration constraints. The resulting reference trajectory $\boldsymbol{\tau}$ contains the wrist trajectory, 3D MANO keypoint trajectories, and object pose trajectory used for subsequent retargeting and reward computation.


\textbf{Robot Embodiment Retargeting.}
We retarget the reconstructed human trajectory to the robot embodiment using a two-stage optimization procedure. We first optimize the arm trajectory to track the reconstructed wrist pose while holding the hand at a nominal configuration. We then jointly optimize the arm and hand to align robot hand keypoints with the reconstructed MANO targets, using larger weights for the
thumb, index finger, and distal keypoints. After retargeting, we obtain a reference trajectory $\tau^r=\{r_t\}_{t=0}^{T}$, which is subsequently used for state initialization and motion-reference conditioning during policy training.

\textbf{Spatial Augmentation.}
Following prior demonstration augmentation pipelines~\cite{li2026momagengeneratingdemonstrationssoft,jiang2025dexmimicgenautomateddatageneration}, we augment each demonstration before retargeting with a shared planar translation and yaw perturbation. For any planar point $\mathbf{x}_{xy}$ from the wrist, MANO target, or object trajectory, we apply
$
\tilde{\mathbf{x}}_{xy}
=
\mathbf{R}_z(\Delta\psi)(\mathbf{x}_{xy}-\mathbf{c}_{xy})
+\mathbf{c}_{xy}
+\Delta\mathbf{p}_{xy},
$
where $\mathbf{c}_{xy}$ is the arm-base position, $\Delta\psi \sim \mathcal{U}(-\psi_{\max},\psi_{\max})$ is the yaw perturbation, and $\Delta\mathbf{p}_{xy}$ is a sampled planar translation within the target workspace.

\vspace{-0.05in}
\subsection{State Expert Training}
\label{subsec:rl}
\vspace{-0.05in}

Kinematic retargeting provides feasible motion references but does not capture the physical interactions required for contact-rich manipulation. We therefore train a closed-loop state-based expert with reinforcement learning in simulation, using $\tau^{r}$ both to sample initial states and provide motion references, while allowing the policy to explore
contact-rich interactions with per-fingertip force feedback. The resulting expert serves as a privileged teacher and provides action supervision for subsequent policy distillation.
We build on PPO~\cite{schulman2017proximalpolicyoptimizationalgorithms} with asymmetric actor-critic~\cite{pinto2017asymmetricactorcriticimagebased}. Domain randomization covers object physics, PD gains, action delay, observation noise, and tactile sensing (Appendix~\ref{app:dr}). The training procedure is summarized in Algorithm~\ref{alg:training}.

\textbf{Tactile Feedback from Simulation.}
The expert receives per-fingertip tactile feedback from simulated contact sensors, represented as five scalar contact-force magnitudes, one for each fingertip. The force signals are averaged over the two recent samples to suppress transient contact spikes. Contact-position channels are retained to maintain a fixed observation layout but are set to zero for the default expert. During training, we randomize tactile latency, apply multiplicative force noise, and randomly drop fingertip signals to improve robustness to real-world sensing imperfections.

\textbf{Observation and Reference.}
The actor observes $\mathbf{s}^{\mathrm{act}}_t$, consisting of proprioception, the next-step retargeted motion reference $\mathbf{r}_{t+1}$, object information, and tactile feedback. Object information includes the reference object pose, fingertip-to-object distances, a BPS geometry encoding~\cite{prokudin2019efficientlearningpointclouds}, and a noisy 6-DoF object-pose estimate. The object-pose observation is randomized with frame-wise noise and episode-wise bias, together with latency and dropouts, to improve robustness to perception errors. The critic additionally receives privileged simulation state, including the ground-truth object state and a five-step future horizon of target object states and fingertip-to-object distances.

 \textbf{Action and Control.}
The policy outputs $\mathbf{a}_t \in \mathbb{R}^{29}$, consisting of 7 arm joint-delta commands and 22 hand joint-position targets. The arm command is applied as
$\mathbf{q}^{\mathrm{arm}}_{t+1}
=
\mathbf{q}^{\mathrm{arm}}_t
+
\alpha \mathbf{a}^{\mathrm{arm}}_t$,
with $\alpha=0.2$\,rad, limiting the per-step arm joint update at 30\,Hz.
 
\textbf{Reward.}
The reward combines wrist and hand tracking, object tracking, contact shaping,
action regularization, and task success:
$r_t = r_t^{\mathrm{track}} + r_t^{\mathrm{object}} + r_t^{\mathrm{contact}}
+ r_t^{\mathrm{action}} + r_t^{\mathrm{success}} + r_t^{\mathrm{collision}}$.
The contact terms include fingertip-force, approach, and no-slip shaping;
detailed reward components and weights are provided in Appendix~\ref{app:reward}.

\vspace{-0.05in}
\subsection{Policy Distillation and Sim-to-Real Transfer}
\vspace{-0.05in}
\label{subsec:transfer}
To enable real-world deployment, we distill the state-based expert into a visual-tactile student policy. The student retains the retargeted motion reference and target object pose for task specification, while replacing the explicit geometry and current-state features with a visual-tactile point-cloud representation.

\textbf{Teacher-Student Distillation.}
We distill the state expert into a single multi-task student using
DAgger~\cite{ross2011reductionimitationlearningstructured}. The student combines a reduced actor-side observation containing task references, proprioception, and fingertip force feedback with a learned point-cloud feature. Training uses aggregated rollout data labeled by the expert, while the expert-student mixing coefficient is gradually decayed over iterations. 

\textbf{Augmented Point-Cloud Observation and Tactile Feedback.} The point cloud contains depth-derived scene points, six robot-hand keypoints, and 25 tactile surface points. Tactile points carry fingertip force magnitudes, and a scalar type indicator distinguishes scene, hand, and tactile points. The point cloud is encoded by a shared PointNet backbone.

\textbf{Real-Time Deployment.} At deployment, the student runs in closed loop conditioned on the motion reference $\mathbf{r}_{t+1}$ together with real-time proprioceptive, depth, and fingertip tactile observations. The policy predicts joint-level commands for the dexterous hand-arm system at 30 Hz. 

\begin{figure*}
    \centering
    \includegraphics[width=\linewidth]{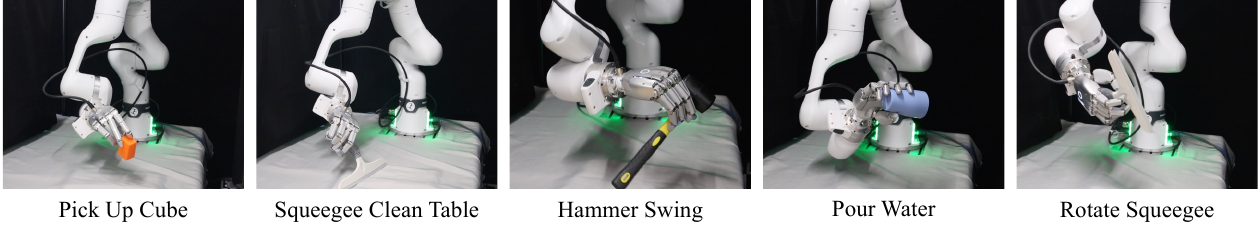}
    \caption{
    Representative real-world visual-tactile dexterous manipulation behaviors across diverse contact-rich tasks.
    }
    \vspace{-0.2in}
    \label{fig:real_snapshots}
\end{figure*}



\section{Experiments}
\label{sec:result}
\vspace{-0.05in}
We structure experimental evaluation around three 
research questions:
\textbf{Q1}: How effectively does \modelname{} learn diverse dexterous manipulation skills in simulation (Sec.~\ref{subsec:Q1})? \textbf{Q2}: How do vision, tactile feedback, and their representations affect manipulation performance (Sec.~\ref{subsec:Q2})? \textbf{Q3}: How well do the learned policies transfer to real-world deployment (Sec.~\ref{subsec:Q3})?
\vspace{-0.05in}
\subsection{Experimental Setup}
\vspace{-0.05in}
We evaluate the proposed framework on a Franka FR3 arm with a Sharpa Wave dexterous hand (29 DoF per arm; 58 DoF in the bimanual configuration).
Visual observations are provided by a fixed RealSense depth camera, while tactile feedback is collected from fingertip tactile sensors. 
We train in IsaacLab~\cite{mittal2025isaaclab} using PPO~\cite{schulman2017proximalpolicyoptimizationalgorithms} with an asymmetric actor-critic architecture~\cite{pinto2017asymmetricactorcriticimagebased} across 4096 parallel environments at 30\,Hz.
All policies are deployed at 30\,Hz. 
The benchmark contains six task categories: pick-up (2 objects, 5 demos: cup and cube), tool use (2 objects, 3 demos: squeegee and hammer), peg insertion (1 object, 2 demos), in-hand rotation (1 object, 2 demos), in-hand translation (1 object, 1 demo), and bimanual handover (1 object, 1 demo). 


\vspace{-0.05in}
\subsection{Q1: Multi-Task Dexterous Manipulation in Simulation}
\vspace{-0.05in}
\label{subsec:Q1}
We first evaluate the state-based expert in simulation against three baselines: a ManipTrans-style imitation baseline~\cite{li2025maniptransefficientdexterousbimanual} adapted to our setup, kinematic retargeting, and a DAPG-style PPO baseline~\cite{rajeswaran2018learningcomplexdexterousmanipulation} that combines demonstrations with reinforcement learning. Table~\ref{tab:multitask_vs_baselines} reports per-category success rates, with averages computed equally across the evaluated categories. \modelname{} achieves an average success rate of \textbf{65.9\%}, compared with 21.9\% for ManipTrans and 5.2\% for kinematic retargeting. On the five single-hand categories evaluated by DAPG, \modelname{} achieves 61.7\% average success compared with 40.6\% for DAPG. The advantage over ManipTrans is particularly pronounced on several contact-rich tasks, including tool use (74.3\% vs.\ 35.0\%), in-hand translation (64.8\% vs.\ 11.5\%), and bimanual handover (86.6\% vs.\ 4.7\%). Qualitatively, we observe that imitation-based policies often lose stable contact during extended interactions, leading to object slip or task failure. In contrast, \modelname{} explores physical interactions with simulated fingertip force feedback during training, enabling closed-loop adaptation throughout manipulation.
\begin{table}[t]
\centering
\small
\setlength{\tabcolsep}{3.0pt}
\caption{Per-category success rate (\%). \modelname{} stage columns report entered-stage success, while baselines report overall success. DAPG is evaluated on the five single-hand categories only.}
\label{tab:multitask_vs_baselines}
\begin{tabular}{l|ccc|c|c|c|c}
\toprule
& \multicolumn{3}{c|}{\textbf{\modelname{} (ours)}}
& \textbf{\modelname{}}
& \textbf{DAPG}
& \textbf{ManipTrans}
& \textbf{Kin.\ Retarget} \\
\cmidrule(lr){2-4}
\textbf{Category}
& \textbf{Reach}
& \textbf{Grasp}
& \textbf{Manip}
& \textbf{Overall}
& \textbf{Overall}
& \textbf{Overall}
& \textbf{Overall} \\
\midrule
Pick Up
& 99.2 & 97.4 & 89.9 & \textbf{89.6} & 66.7 & 1.5 & 1.8 \\
Peg Insertion$^{\ast}$
& 56.8 & 43.7 & 43.3 & 43.1 & \textbf{52.8} & --- & 0.0 \\
Tool Use
& 86.9 & 82.3 & 74.6 & \textbf{74.3} & 70.1 & 35.0 & 24.1 \\
In-hand Rotation
& --- & --- & 36.8 & 36.8 & 0.7 & \textbf{56.9} & 0.0 \\
In-hand Translation
& --- & --- & 64.8 & \textbf{64.8} & 12.5 & 11.5 & 0.0 \\
Bimanual Handover
& 99.0 & 88.8 & 87.9 & \textbf{86.6} & --- & 4.7 & --- \\
\midrule
\textbf{Average}
& \textbf{85.5}
& \textbf{78.0}
& \textbf{66.2}
& \textbf{65.9}
& \textbf{40.6}$^{\dagger}$
& \textbf{21.9}
& \textbf{5.2} \\
\bottomrule
\end{tabular}

\par\smallskip
\raggedright
{\footnotesize
$^{\ast}$~Peg insertion uses a $<\!1$\,cm success criterion.
$^{\dagger}$~DAPG average is computed over the five evaluated single-hand categories.
}
\vspace{-0.1in}
\end{table}

\vspace{-0.05in}
\subsection{Q2: The Role of Scene and Tactile Representations}
\vspace{-0.05in}
\label{subsec:Q2}
\paragraph{Simulation Ablation.}
We investigate how visual geometry and tactile feedback contribute to the deployable student policy, and compare alternative visual-tactile representations.  We report two success metrics: a \emph{strict} metric requiring the final object position to be within 3\,cm of the target, and a \emph{relaxed} metric using a 5\,cm threshold. For rotation tasks, a 30$^\circ$ orientation threshold is applied.
\begin{figure}[!t]
\centering
\includegraphics[width=\linewidth]{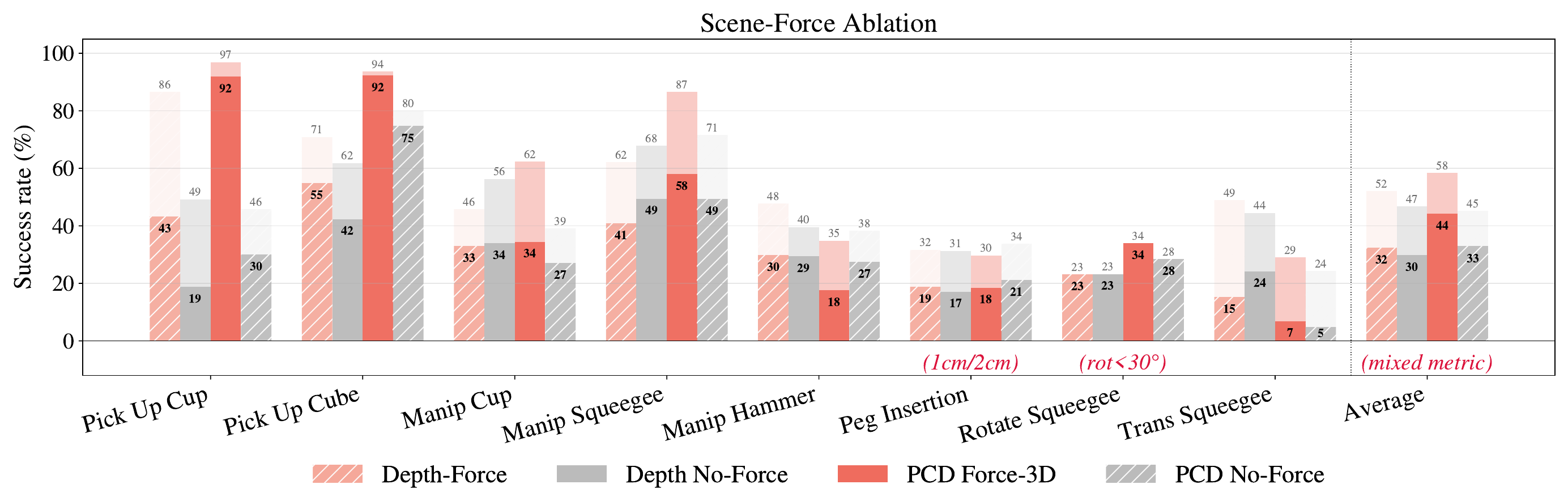}

\caption{\textbf{Scene Representation Ablation.}
Per-task success rates related to scene representations (depth vs.\ point cloud) and contact-force usage (with vs.\ without force).
Dark and light bars indicate the strict and relaxed criteria, respectively.
Point clouds with local 3D force achieve the best average performance, and removing force generally reduces performance, especially for point-cloud. 
}
\label{fig:scene_modality_ablation}
\vspace{-0.1in}
\end{figure}



In simulation, point-cloud scene observations with tactile feedback achieve 44\% strict and 58\% relaxed success, compared with 32\% strict success using depth-based observations. Removing tactile feedback from the point-cloud policy further reduces relaxed success from 58\% to 45\%, suggesting that both explicit 3D geometry and tactile feedback improve manipulation performance. We additionally compare scalar force, binary contact, and 3D force representations, and observe similar performance across these tactile encodings. We therefore adopt scalar force magnitudes for deployment to maintain consistency with real-world tactile sensing.

\paragraph{Real-World Modality Ablation.}
We further evaluate the contribution of vision and tactile feedback through controlled real-world ablations on cube picking. All variants are distilled from the same teacher. The full visual-tactile policy achieves 28/30 success, while removing vision or tactile feedback reduces success to 14/30 and 11/30, respectively; using proprioception alone further reduces success to 8/30. Common failures include missing the object during grasp acquisition and losing the object after contact is established. These results confirm that both visual and tactile observations are important for robust real-world deployment.

\begin{figure}[t]
\centering
\includegraphics[width=0.95\linewidth]{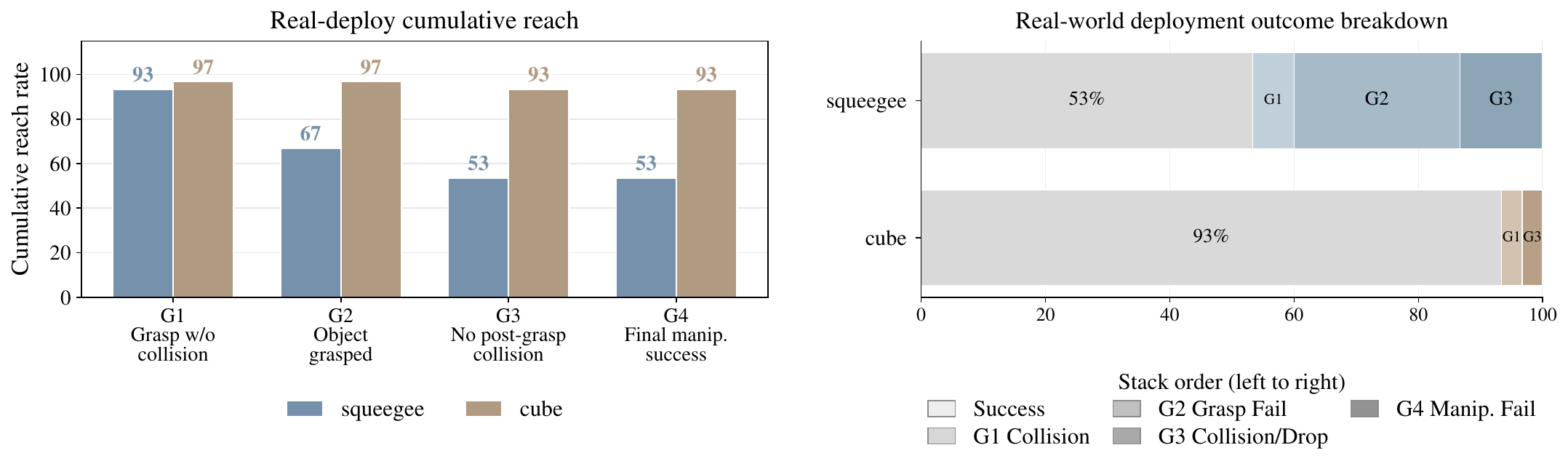}
\caption{\textbf{Real-world deployment on the squeegee and cube tasks}  (30 trials per task, manually annotated). (a) Cumulative success across execution stages: G1, grasp without collision; G2, object grasped; G3, no post-grasp collision or drop; and G4, final manipulation success. The cube policy maintains high success throughout all stages (93$\%$ final success), while the squeegee task shows the largest performance drop during grasp acquisition and reaches 53$\%$ final success. (b) Outcome breakdown over all trials, showing that squeegee failures are dominated by grasp failures and post-grasp collision/drop events, whereas cube failures are rare and arise primarily from collision-related events.}
\vspace{-0.05in}
\label{fig:real_deploy}
\end{figure}

\begin{table}[t]
\centering
\small

\begin{minipage}[t]{0.48\linewidth}
\centering
\captionof{table}{
Real-world observation ablation on cube picking.
}
\label{tab:realworld_obs_ablation}

\setlength{\tabcolsep}{8pt}
\renewcommand{\arraystretch}{1.08}
\begin{tabular}{lc}
\toprule
\textbf{Observation} & \textbf{Success} \\
\midrule
Full Visual-Tactile & \textbf{28/30 (93.3\%)} \\
No Vision           & 14/30 (46.7\%) \\
No Tactile          & 11/30 (36.7\%) \\
Proprioception Only & 8/30 (26.7\%) \\
\bottomrule
\end{tabular}
\end{minipage}
\hfill
\begin{minipage}[t]{0.48\linewidth}
\centering
\captionof{table}{
Zero-shot object generalization on cube picking.
}
\label{tab:realworld_object_generalization}

\setlength{\tabcolsep}{8pt}
\renewcommand{\arraystretch}{1.08}
\begin{tabular}{lc}
\toprule
\textbf{Object} & \textbf{Success} \\
\midrule
Training Cube        & \textbf{28/30 (93.3\%)} \\
Thin Cube (unseen)   & 23/30 (76.7\%) \\
Square Cube (unseen) & 8/30 (26.7\%) \\
Big Duck (unseen)    & 8/30 (26.7\%) \\
Small Duck (unseen)  & 7/30 (23.3\%) \\
\bottomrule
\end{tabular}
\end{minipage}

\vspace{-2mm}
\end{table}
\vspace{-0.05in}
\subsection{Q3: Real-World Deployment and Generalization}
\vspace{-0.05in}
\label{subsec:Q3}
\paragraph{Real-World Deployment.}
We evaluate the distilled visual-tactile policy on real-world dexterous
manipulation without additional fine-tuning. Across four tasks, the policy
achieves 28/30 success on cube picking, 24/30 on cup pouring, 22/30 on cup
lifting, and 16/30 on squeegee manipulation. These results demonstrate
zero-shot sim-to-real transfer across grasping and contact-rich tool
interaction. The lower success rate on squeegee manipulation reflects the
difficulty of maintaining stable grasp and sustained contact over a longer
interaction horizon.
\paragraph{Object Generalization.}
We further evaluate zero-shot object generalization on the cube-picking task.
Without additional training, the policy achieves 23/30 success on an unseen
thin cube, compared with 28/30 on the training cube. Larger geometry changes
reduce success to 7--8/30 on an unseen square cube and two toy ducks,
demonstrating transfer to unseen object geometries while also revealing the
current limits of object generalization.

Together, these results show that the distilled policy transfers across diverse real-world interactions and exhibits zero-shot generalization beyond the training object.

\vspace{-0.1in}
\section{Conclusion}
\vspace{-0.1in}
\label{sec:conclusion}
We presented \modelname{}, a framework for learning visual-tactile dexterous manipulation policies from monocular human video demonstrations. Our key insight is to reconstruct human demonstrations as physical interactions in simulation, where contact dynamics provide tactile supervision that is unavailable in the original videos. By combining human-video-derived motion references with tactile-aware reinforcement learning, \modelname{} transforms passive visual demonstrations into closed-loop interaction policies and enables zero-shot sim-to-real transfer to a 29-DoF dexterous hand-arm platform. Across diverse grasping and contact-rich tool-use tasks, the learned policies substantially outperform imitation and retargeting baselines, transfer to real-world execution without additional fine-tuning, and exhibit zero-shot generalization to unseen object geometries. Real-world sensory ablations further confirm the importance of both visual and tactile feedback for robust deployment. Together, these results highlight simulated physical interaction as an effective bridge between human demonstrations and deployable dexterous robot policies, providing a path toward scaling robot skill learning from abundant human video data.

\vspace{-0.1in}
\section{Limitations}
\vspace{-0.1in}

Robust long-horizon interaction and broad generalization remain challenging.
Performance is sensitive to perception errors, controller mismatch, and large
geometry or category shifts, while the deployed policy still relies on
retargeted motion references. Scaling to broader skills will require more
diverse demonstrations, simulated interactions, and more flexible task
conditioning.

Our tactile representation is limited to per-fingertip force magnitudes and
does not capture richer signals such as pressure, shear, slip, or contact
patches. Higher-resolution tactile sensing could support more precise contact
reasoning and force regulation.

\clearpage
\acknowledgments{This work is supported in part by the Dushi Program. The views and opinions expressed in this work are solely those of the authors.}


\bibliography{example}  

\clearpage
\appendix

\section*{Appendix}

\section{Supplementary Videos and Demos}
We provide additional qualitative results, including full-length rollout videos and diverse task demonstrations, on our project website: \href{https://dexx-code.github.io/dexx-code/}{https://dexx-code.github.io}. We encourage readers to visit the website for a more comprehensive view of our system's capabilities.






\section{Analysis of Sim-to-Real Transfer}

We further collect rollouts in simulation and compare them with real-world rollouts to analyze whether the learned contact behaviors and motion trajectories transfer across domains.

\paragraph{Fingertip force patterns.}
For each rollout, we record the contact force at all five fingertips and visualize their distributions over manipulation time.

\begin{figure}[H]
    \centering
    \includegraphics[width=\linewidth]{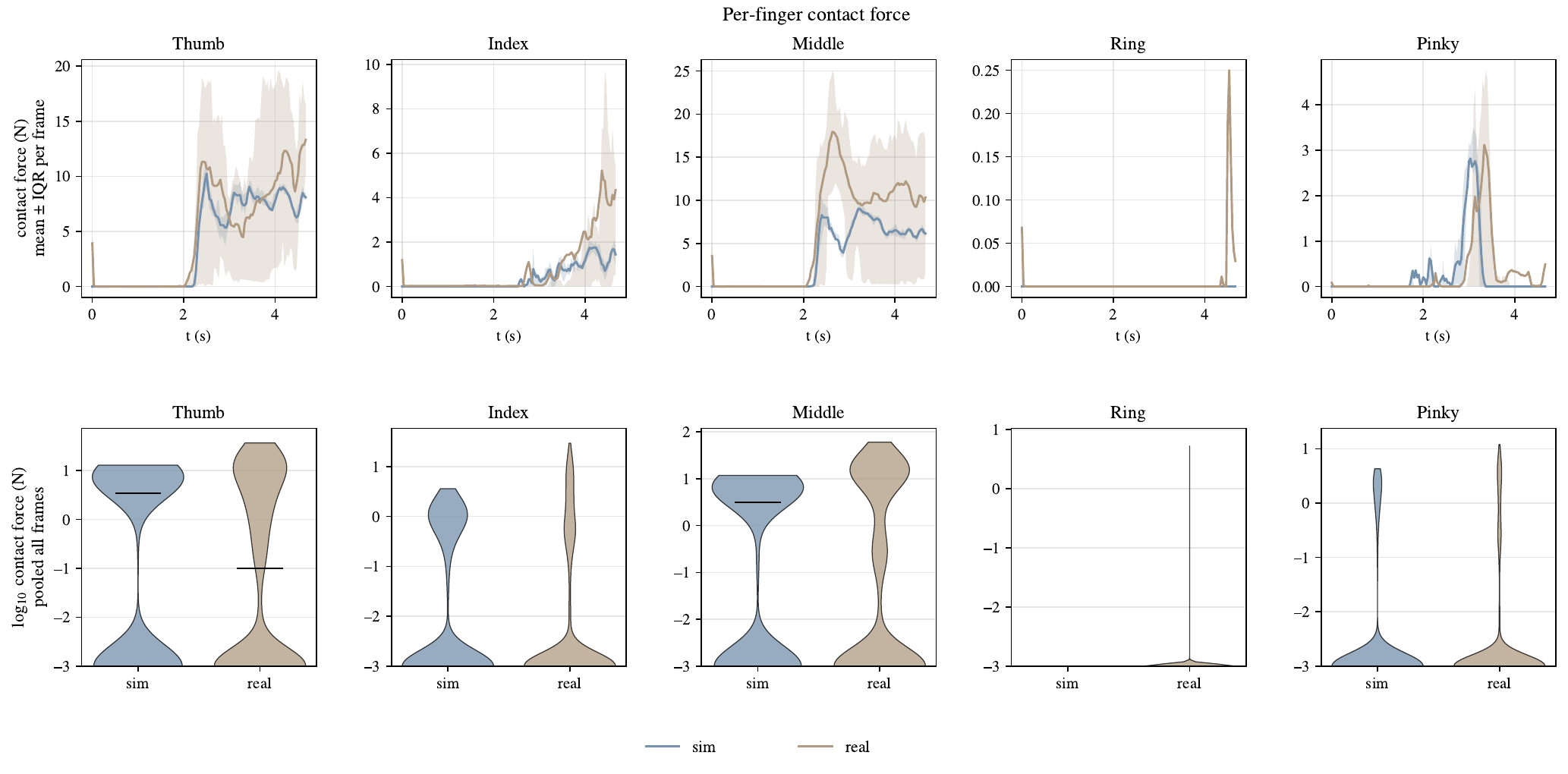}
    \caption{
    Comparison of fingertip force distributions between simulation and real-world rollouts during a cube-picking task. 
    We visualize the temporal evolution of contact forces across five fingertips. 
    The policy exhibits consistent force patterns in both domains, including contact timing, force allocation, and force modulation during grasp stabilization, indicating successful transfer of contact-rich behaviors from simulation to the real robot.
    }
    \label{fig:force_analysis}
\end{figure}
As shown in Figure~\ref{fig:force_analysis}, the policy exhibits remarkably similar force patterns in simulation and reality. The onset of contacts, force magnitudes, and force redistribution among fingers follow consistent trends throughout the manipulation process. Although real-world measurements are noisier due to sensing uncertainty and hardware imperfections, the overall structure of the force profiles remains largely preserved.
This observation suggests that the policy transfers not only the motion trajectories but also the contact strategies learned in simulation. The consistency of fingertip force distributions indicates that the policy has learned physically meaningful interaction behaviors that generalize to real-world execution.

\paragraph{Wrist trajectory analysis.}
To quantify the spatial fidelity of sim-to-real transfer, we compare wrist trajectories generated by the policy in simulation with those observed during real-world deployment.

\begin{figure}[h]
    \centering
    \includegraphics[width=\linewidth]{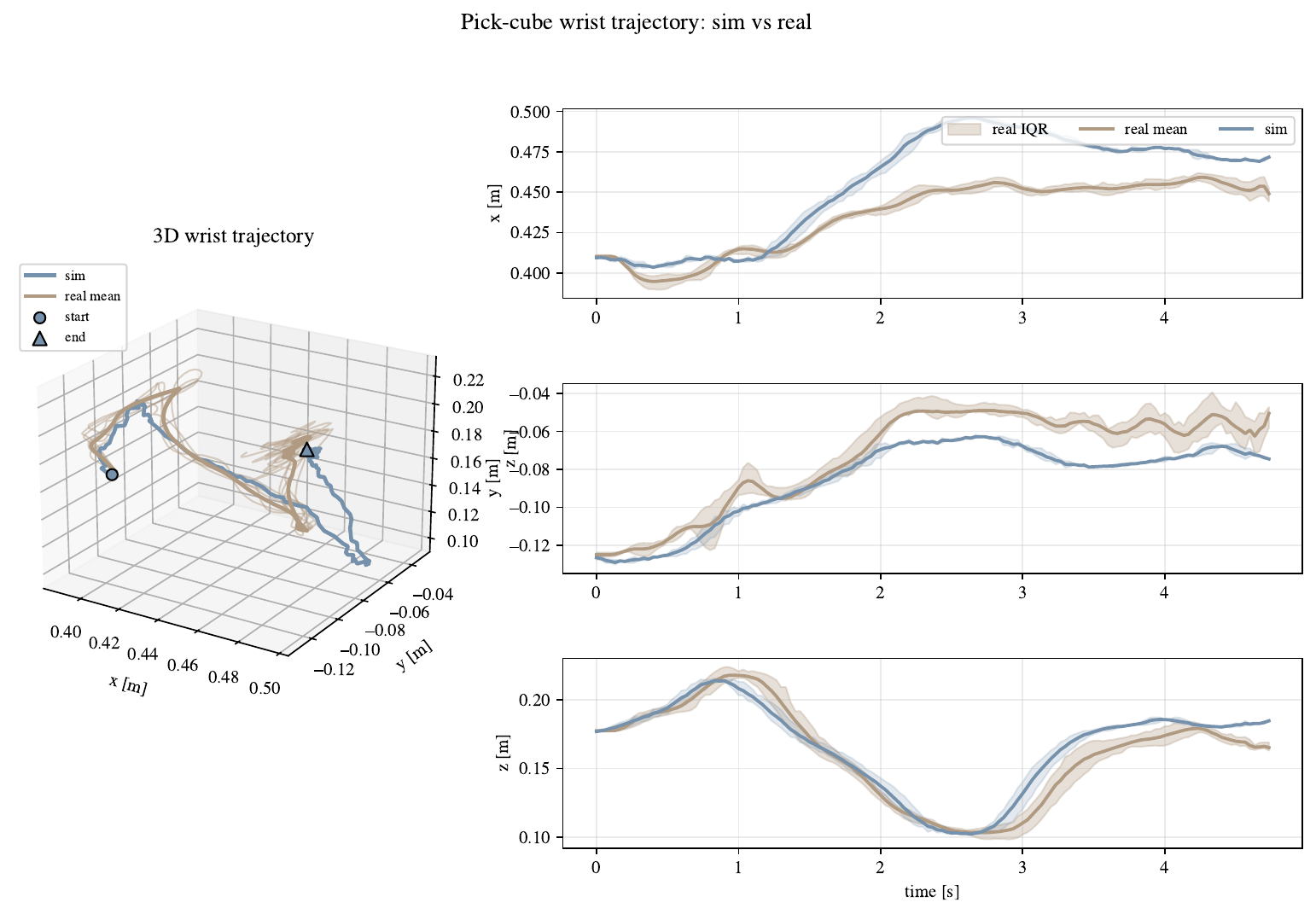}
    \caption{
    We compare the simulated rollout (blue) with real-world rollouts (brown).
For visualization, the 3D trajectory plot shows five randomly sampled
real-world rollouts together with the simulated trajectory. The coordinate-wise
plots show the mean and interquartile range (IQR) of the collected real-world
rollouts, with the simulated rollout overlaid for comparison.
Across episodes, the real-world trajectories remain highly consistent
(std.\ $0.46$--$0.82$\,cm) and exhibit a mean sim-to-real tracking error of
$2.83$\,cm. The dominant discrepancy occurs along the $x$-axis under payload,
consistent with the steady-state compliance offset of the arm impedance
controller.
    }
    \label{fig:wrist_traj}
\end{figure}
As shown in Figure~\ref{fig:wrist_traj}, the real-world trajectories closely follow the simulated rollout, with a mean sim-to-real tracking error of $2.83$\,cm. The cross-episode standard deviation of the real-world trajectories remains small ($0.46$--$0.82$\,cm), indicating high repeatability across trials.  Note that real-world deployment introduces additional sources of spatial uncertainty not present in simulation, including up to $1$\,cm of object placement variation across episodes and camera extrinsic calibration error that affects the estimated object pose in the point cloud observation.

The sim-to-real discrepancy exhibits structured patterns across spatial axes and task phases. The dominant deviation occurs along the $x$-axis, which corresponds to the forward load-bearing direction, while the $y$- and $z$-trajectories remain more closely aligned. Temporally, the discrepancy increases from $1.1$\,cm during approach to $2.7$\,cm during grasp and $3.8$\,cm during lift. This pattern is consistent with a steady-state compliance offset of the arm impedance controller under increasing payload,
together with object-placement variation and camera-calibration error. Along the $z$-axis, the real robot consistently lifts approximately $5.7$\,mm higher than in simulation, suggesting a systematic vertical offset during execution.

These observations suggest that the residual sim-to-real gap in wrist tracking is dominated by systematic arm-compliance effects, object-placement variation, and camera-calibration error, rather than large differences in the learned motion itself. The gap could be further reduced through improved arm-dynamics calibration and more accurate camera extrinsic calibration.

\section{Hand-Object Reconstruction Details}
\label{app:reconstruction}

Our reconstruction pipeline consists of three stages: object pose estimation, hand pose estimation, and joint hand-object optimization.

\textbf{Object pose estimation.}
Object meshes are prepared offline using either 3D scanning or single-image
reconstruction; this step is independent of the demonstration recording.
Given a monocular demonstration, we estimate the per-frame 6-DoF object pose
using FoundationPose~\cite{wen2024foundationposeunified6dpose}, represented by
position $\mathbf{p}^O_t \in \mathbb{R}^3$ and rotation
$\mathbf{R}^O_t \in \mathrm{SO}(3)$.

\textbf{Hand pose estimation.}
We use WiLoR~\cite{potamias2025wilorendtoend3dhand} to obtain an initial
per-frame hand estimate from the monocular RGB video. We then refine the hand
using the MANO parametric model. For each frame, we optimize the MANO pose
$\theta^H_t \in \mathbb{R}^{45}$, root translation
$\mathbf{t}_t \in \mathbb{R}^3$, and global rotation
$\mathbf{R}_t \in \mathrm{SO}(3)$ using the detected hand keypoints:
\begin{equation}
    \mathcal{L}_{\mathrm{hand}}
    =
    \sum_{k=1}^{21}
    \left\|
    \pi\!\left(
    \mathbf{J}_k(\theta^H_t,\beta,\mathbf{t}_t,\mathbf{R}_t)
    \right)
    -
    \mathbf{j}^{k}_t
    \right\|^2 ,
\end{equation}
where $\pi$ denotes the projection function of the monocular camera,
$\mathbf{J}_k$ is the $k$-th MANO keypoint, $\beta$ is a fixed hand-shape
parameter, and $\mathbf{j}^{k}_t$ is the detected 2D keypoint.

\textbf{Temporal refinement.}
After per-frame estimation, we jointly refine the hand trajectory over the
full sequence with an additional temporal smoothness objective:
\begin{equation}
    \mathcal{L}_{\mathrm{batch}}
    =
    \sum_t \mathcal{L}_{\mathrm{hand},t}
    +
    \lambda_{\mathrm{cont}}
    \sum_t
    \left\|
    \mathbf{V}_{t+1}-\mathbf{V}_t
    \right\|^2 ,
\end{equation}
where $\mathbf{V}_t$ denotes the MANO mesh vertices at frame $t$.
The sequence is optimized using AdamW for 5000 iterations with learning rate
$10^{-3}$ and a step-decay factor of $0.5$ every 1000 iterations.

\textbf{Joint hand-object optimization.}
The independently estimated hand and object trajectories may contain
interpenetration and temporal jitter. We jointly refine them using
\begin{equation}
    \mathcal{L}_{\mathrm{joint}}
    =
    w_{\mathrm{sdf}} \mathcal{L}_{\mathrm{sdf}}
    +
    w_{\mathrm{reg,h}} \mathcal{L}_{\mathrm{reg,h}}
    +
    w_{\mathrm{reg,o}} \mathcal{L}_{\mathrm{reg,o}}
    +
    w_{\mathrm{sm,h}} \mathcal{L}_{\mathrm{sm,h}}
    +
    w_{\mathrm{sm,o}} \mathcal{L}_{\mathrm{sm,o}} .
\end{equation}

The penetration term $\mathcal{L}_{\mathrm{sdf}}$ penalizes hand vertices
inside the object surface. The registration terms
$\mathcal{L}_{\mathrm{reg,h}}$ and $\mathcal{L}_{\mathrm{reg,o}}$ constrain
the refined trajectories to remain close to their initial estimates, while
$\mathcal{L}_{\mathrm{sm,h}}$ and $\mathcal{L}_{\mathrm{sm,o}}$ regularize
temporal variation in the hand and object trajectories. Optimization is
performed in two stages: we first optimize registration and smoothness terms,
then activate the penetration loss to resolve hand-object interpenetration.

\section{Retargeting Details}
\label{app:retarget}

Given a reconstructed demonstration with wrist poses
$(\mathbf{p}^W_t,\mathbf{R}^W_t)$, MANO hand keypoints
$\{\mathbf{x}^j_t\}$, and object poses
$(\mathbf{p}^O_t,\mathbf{R}^O_t)$, we retarget the human motion to
the Franka FR3 and Sharpa HA4 embodiment through a two-stage optimization
procedure. All targets are first transformed into the simulation workspace
using a fixed coordinate transformation. We then optimize the arm and hand
joint trajectories to track the reconstructed wrist and hand motion while
respecting the kinematic limits of the physical robot.

\textbf{Spatial augmentation.}
Before optimization, we generate spatially augmented variants of each
demonstration by applying a shared planar translation and yaw rotation to
the wrist, MANO keypoints, and object trajectory. For a sampled yaw
$\Delta\psi$, planar points are rotated around the robot arm base
$\mathbf{c}_{xy}$ as
\begin{equation}
    \bar{\mathbf{x}}_{xy}
    =
    \mathbf{R}_z(\Delta\psi)
    (\mathbf{x}_{xy}-\mathbf{c}_{xy})
    +
    \mathbf{c}_{xy}.
\end{equation}
The corresponding wrist and object orientations are left-multiplied by
$\mathbf{R}_z(\Delta\psi)$.

We first align the initial object position with a predefined workspace
anchor $\mathbf{p}^{\mathrm{anchor}}_{xy}$ and then apply an additional
sampled translation
$\Delta\mathbf{p}_{xy}\sim
\mathcal{U}([-5\,\mathrm{cm},5\,\mathrm{cm}]^2)$.
The resulting shared translation is
\begin{equation}
    \mathbf{d}_{xy}
    =
    \mathbf{p}^{\mathrm{anchor}}_{xy}
    +
    \Delta\mathbf{p}_{xy}
    -
    \bar{\mathbf{p}}^{O}_{0,xy},
\end{equation}
and all positional targets are translated by
\begin{equation}
    \tilde{\mathbf{x}}_{xy}
    =
    \bar{\mathbf{x}}_{xy}
    +
    \mathbf{d}_{xy}.
\end{equation}
We sample
$\Delta\psi\sim\mathcal{U}(-10^\circ,10^\circ)$.
Applying the same transformation to the wrist, hand, and object preserves
their relative interaction geometry.

\textbf{Stage 1: arm-only wrist optimization.}
We first optimize the arm trajectory to track the reconstructed wrist pose.
The hand is held at a nominal configuration, and the first arm joint
(base yaw) is fixed at its default value. Only arm joints 2--7 are optimized.
At each iteration, the optimized joints are clamped to their URDF joint
limits.

Let
$\mathbf{p}^{\mathrm{ee}}_t(\mathbf{q}^{\mathrm{arm}}_t)$ and
$\mathbf{R}^{\mathrm{ee}}_t(\mathbf{q}^{\mathrm{arm}}_t)$ denote the
end-effector position and orientation obtained through forward kinematics.
We minimize
\begin{equation}
    \mathcal{L}_{\mathrm{stage1}}
    =
    0.5\,\mathcal{L}_{\mathrm{pos}}
    +
    0.25\,\mathcal{L}_{\mathrm{rot}}
    +
    10^{-3}\mathcal{L}^{\mathrm{arm}}_{\mathrm{vel}},
\end{equation}
where
\begin{equation}
    \mathcal{L}_{\mathrm{pos}}
    =
    \frac{1}{T}
    \sum_t
    \left\|
    \mathbf{p}^{\mathrm{ee}}_t
    -
    \tilde{\mathbf{p}}^W_t
    \right\|_2 ,
\end{equation}
and
\begin{equation}
    \mathcal{L}_{\mathrm{rot}}
    =
    \frac{1}{T}
    \sum_t
    d_{\mathrm{geo}}
    \left(
    \mathbf{R}^{\mathrm{ee}}_t,
    \tilde{\mathbf{R}}^W_t
    \right).
\end{equation}
Here $d_{\mathrm{geo}}$ denotes the geodesic rotation distance.
The temporal regularizer
$\mathcal{L}^{\mathrm{arm}}_{\mathrm{vel}}$
penalizes squared arm joint velocities computed over the 30\,Hz
demonstration trajectory.

\textbf{Stage 2: joint arm-hand optimization.}
Starting from the Stage-1 solution, we jointly optimize the complete hand
configuration and arm trajectory. All 22 Sharpa hand joints are optimized,
arm joints 2--7 remain trainable, and the first arm joint is released to
provide additional yaw reach.

For each robot hand body, forward kinematics gives a 3D keypoint
$\mathrm{FK}_j(\mathbf{q}_t)$. The corresponding wrist and MANO targets are
grouped into $\tilde{\mathbf{x}}^j_t$. We minimize the weighted keypoint
tracking objective
\begin{equation}
    \mathcal{L}_{\mathrm{hand}}
    =
    \frac{1}{TK}
    \sum_{t,j}
    w_j
    \left\|
    \mathrm{FK}_j(\mathbf{q}_t)
    -
    \tilde{\mathbf{x}}^j_t
    \right\|_2 ,
\end{equation}
where $K$ is the number of tracked hand-body keypoints.

The tracking weight is factorized into a per-finger weight and a
within-finger level weight,
\begin{equation}
    w_j = w^{\mathrm{finger}}_j
          w^{\mathrm{level}}_j .
\end{equation}
We use larger weights for the thumb and index finger and emphasize distal
keypoints, particularly the fingertips.

The full Stage-2 objective is
\begin{equation}
\begin{aligned}
    \mathcal{L}_{\mathrm{stage2}}
    ={}&
    \mathcal{L}_{\mathrm{hand}}
    +
    0.05\,\mathcal{L}_{\mathrm{pos}}
    +
    0.05\,\mathcal{L}_{\mathrm{rot}} \\
    &+
    10^{-3}\mathcal{L}^{\mathrm{arm}}_{\mathrm{vel}}
    +
    10^{-4}\mathcal{L}^{\mathrm{hand}}_{\mathrm{vel}},
\end{aligned}
\end{equation}
where the last two terms penalize squared frame-to-frame joint velocities
for the arm and hand, respectively.

\textbf{Hand-keypoint tracking weights.}
The per-finger base weights and within-finger level scales used in the
retargeting objective are shown in Table~\ref{tab:retarget_weights}.

\begin{table}[h]
\centering
\small
\caption{Tracking weights used for hand-keypoint retargeting. The final
weight of a keypoint is the product of its finger weight and level scale.}
\label{tab:retarget_weights}
\begin{tabular}{lccccc}
\toprule
 & Thumb & Index & Middle & Ring & Pinky \\
\midrule
Finger weight
& 25 & 15 & 10 & 7 & 5 \\
\bottomrule
\end{tabular}

\vspace{1mm}

\begin{tabular}{lcccc}
\toprule
Level & Tip & Distal & Intermediate & Proximal \\
\midrule
Scale
& 1.0 & 0.6 & 0.4 & 0.3 \\
\bottomrule
\end{tabular}
\end{table}

\textbf{Hardware constraints and reachability filtering.}
During Stage 2, the Sharpa hand joint limits are tightened to the empirically
measured reachable range of the physical hand, which can be more restrictive
than the URDF limits. The optimized hand trajectory is clamped to the same
range before being stored, ensuring that retargeting and real-world
deployment use a consistent feasible action space.

After optimization, we compute the mean end-effector position tracking error
over the trajectory. Augmented variants whose mean error exceeds
$8\,\mathrm{cm}$ are marked unreachable and excluded from policy training.
The resulting retargeted trajectory stores the optimized wrist pose, arm and
hand joint configurations, and robot hand-body positions used to construct
the motion references for subsequent policy training.

\begin{figure}[t]
    \centering
    \includegraphics[width=0.92\linewidth]{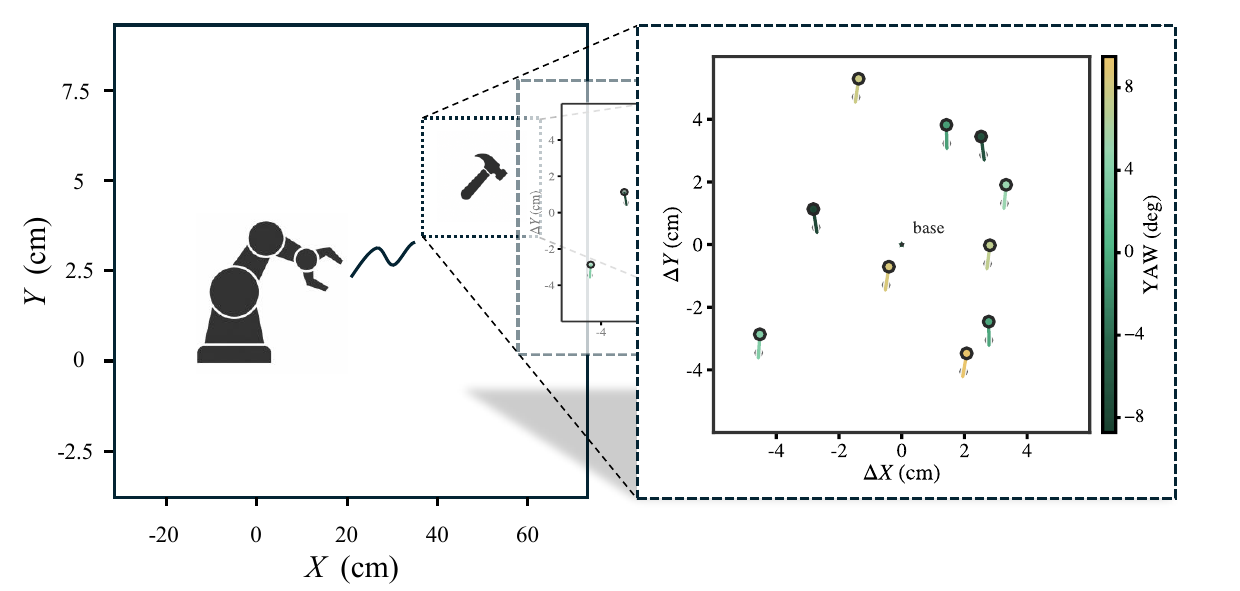}
    \caption{
    Spatial augmentation before retargeting. We sample planar offsets around the predefined target position and apply yaw perturbations about the robot arm base. Points visualize the sampled translation offsets, while oriented markers and color encode the corresponding yaw angle.
    }
    \label{fig:spatial_augmentation}
\end{figure}

\section{Training Algorithms}
\label{app:algorithms}

This section provides the algorithmic specification of state-expert training
(Algorithm~\ref{alg:training}) and teacher-student distillation
(Algorithm~\ref{alg:distill}), followed by the corresponding observation,
reward, and randomization details.

\begin{figure*}[t]
\centering
\begin{minipage}[t]{0.48\textwidth}
\begin{algorithm}[H]
\caption{State Expert Training}
\label{alg:training}
\small
\begin{algorithmic}[1]
\Require Retargeted demonstrations $\mathcal{D}$
\Require Simulation environment with domain randomization
\State Initialize actor $\pi_\theta$ and critic $V_\phi$
\For{each PPO iteration}
    \State Sample $4096$ parallel environments
    \State Assign each environment a demonstration from $\mathcal{D}$
    \State Randomize physics, actuation, and sensing parameters
    \For{each timestep $t$}
        \State Construct actor observation
        $\mathbf{s}^{\mathrm{act}}_t\in\mathbb{R}^{557}$
        \State Construct privileged state
        $\mathbf{s}^{\mathrm{priv}}_t\in\mathbb{R}^{148}$
        \State Construct critic input
        $[\mathbf{s}^{\mathrm{act}}_t,\mathbf{s}^{\mathrm{priv}}_t]$
        \State Sample raw action
        $\mathbf{a}^{\mathrm{raw}}_t
        \sim\pi_\theta(\cdot\mid\mathbf{s}^{\mathrm{act}}_t)$
        \State Clip $\mathbf{a}^{\mathrm{raw}}_t$ to $[-1,1]$
        \State Apply per-environment action delay of $0$--$3$ steps
        \State Split action into $7$ arm and $22$ hand commands
        \State Set arm target
        $\tilde{\mathbf{q}}^{\mathrm{arm}}_t
        =\mathbf{q}^{\mathrm{arm}}_t
        +0.2\,\mathbf{a}^{\mathrm{arm}}_t$
        \State Map hand action to absolute joint-position targets
        \State Low-pass filter arm/hand targets with $0.15/0.4$
        \State Saturate targets to joint limits
        \State Step simulator and compute reward $r_t$
    \EndFor
    \State Compute advantages with GAE
    \State Update $\pi_\theta$ and $V_\phi$ using PPO
\EndFor
\State \Return Teacher policy $\pi_{\mathrm{teacher}}=\pi_\theta$
\end{algorithmic}
\end{algorithm}
\end{minipage}%
\hfill
\begin{minipage}[t]{0.48\textwidth}
\begin{algorithm}[H]
\caption{Teacher-Student Distillation}
\label{alg:distill}
\small
\begin{algorithmic}[1]
\Require Trained teacher $\pi_{\mathrm{teacher}}$
\Require Point-cloud simulation environment
\Require Mixing coefficient $\beta_0=1$
\State Initialize student $\pi_{\mathrm{student}}$ and buffer $\mathcal{B}$
\For{DAgger iteration $k=0,\ldots,K-1$}
    \State $\beta_k\gets\beta_0(0.85)^k$
    \For{each rollout timestep $t$}
        \State Construct teacher observation
        $\mathbf{s}^{\mathrm{act}}_t\in\mathbb{R}^{557}$
        \State Remove BPS, tip-distance, and noisy object-pose blocks
        \Statex \hspace{1.5em}
        $\rightarrow\mathbf{s}^{\mathrm{stu}}_t\in\mathbb{R}^{417}$
        \State Construct visual-tactile point cloud
        $\mathbf{P}_t\in\mathbb{R}^{1055\times5}$
        \State $\mathbf{z}_t\gets
        \phi_{\mathrm{PC}}(\mathbf{P}_t)\in\mathbb{R}^{64}$
        \State $\mathbf{a}^{\mathrm{tea}}_t
        \gets\pi_{\mathrm{teacher}}(\mathbf{s}^{\mathrm{act}}_t)$
        \State $\mathbf{a}^{\mathrm{stu}}_t
        \gets\pi_{\mathrm{student}}
        ([\mathbf{s}^{\mathrm{stu}}_t,\mathbf{z}_t])$
        \State Execute
        $\mathbf{a}^{\mathrm{exec}}_t
        =\beta_k\mathbf{a}^{\mathrm{tea}}_t
        +(1-\beta_k)\mathbf{a}^{\mathrm{stu}}_t$
        \State Store student observation and
        $\mathbf{a}^{\mathrm{tea}}_t$ in $\mathcal{B}$
    \EndFor
    \For{$8$ training epochs}
        \State Sample minibatch from $\mathcal{B}$
        \State Minimize
        $\mathcal{L}_{\mathrm{BC}}
        =\|\mathbf{a}^{\mathrm{stu}}
        -\mathbf{a}^{\mathrm{tea}}\|_2^2$
    \EndFor
\EndFor
\State \Return Deployable policy $\pi_{\mathrm{student}}$
\end{algorithmic}
\end{algorithm}
\end{minipage}
\end{figure*}

\paragraph{State expert training.}
Algorithm~\ref{alg:training} summarizes state-expert training with PPO and
an asymmetric actor-critic architecture. The actor receives a
557-dimensional observation comprising proprioception (79), retargeted wrist
and hand motion references (311), the demonstration target object pose (7),
reference fingertip-to-object distances (5), a static BPS encoding of the
object mesh (128), tactile-related channels (20), and a noisy current-object
pose estimate (7). The tactile block contains five active fingertip-force
magnitudes and 15 reserved contact-position channels that are disabled and
set to zero in the default configuration.

The critic concatenates the actor observation with a 148-dimensional
privileged state, yielding a 705-dimensional critic input. The privileged
state includes hand joint velocities, ground-truth object state and dynamics,
a five-step horizon of future target object states and fingertip-to-object
distances, and current object-to-target relation features.

The policy outputs 29 actions, consisting of seven arm commands and 22 hand
commands. Arm actions are interpreted as joint-position deltas with scale
$0.2$, while hand actions specify absolute joint-position targets after
mapping from the normalized action range to the hand joint limits. The raw
29-dimensional action is first clipped and subjected to a randomized delay
of $0$--$3$ control steps before being split into arm and hand commands.
Separate low-pass filters with coefficients $0.15$ and $0.4$ are then applied
to the arm and hand targets, respectively, before saturation to the joint
limits.

\paragraph{Distillation.}
Algorithm~\ref{alg:distill} describes teacher-student distillation using
DAgger. The student receives a reduced version of the teacher observation.
Specifically, we remove the 128-dimensional BPS object-shape encoding, the
5-dimensional reference fingertip-to-object distance features, and the
7-dimensional noisy current-object pose estimate, reducing the
557-dimensional teacher observation to 417 dimensions. The retargeted motion
reference, including the target object pose, is retained for task
conditioning.

To provide scene geometry and current object information from visual
observations, the student additionally receives a visual-tactile point-cloud
representation. The point cloud contains 1024 scene points, six robot-hand
keypoints (the wrist and five fingertips), and 25 tactile surface points
(five per fingertip). Each point carries its 3D position, a scalar point-type
indicator, and a scalar tactile-force channel. The force channel is zero for
scene and hand points and carries the corresponding fingertip-force magnitude
for tactile points. A shared PointNet encoder with masked max-pooling maps
the resulting $1055\times5$ point cloud to a 64-dimensional feature.

The student therefore receives a $417+64=481$ dimensional input and predicts
the same 29-dimensional action as the teacher. During DAgger rollouts, the
teacher and student actions are combined through the convex mixture
\begin{equation}
    \mathbf{a}^{\mathrm{exec}}_t
    =
    \beta_k\,\mathbf{a}^{\mathrm{tea}}_t
    +
    (1-\beta_k)\,\mathbf{a}^{\mathrm{stu}}_t .
\end{equation}
The mixing coefficient starts from $1.0$ and is geometrically decayed by a
factor of $0.85$ per iteration. We train for 30 DAgger iterations with 4096
rollout steps per iteration. The aggregated replay buffer, with a capacity
of $2\times10^5$ transitions, is retained across iterations, and the student
is trained for eight epochs per iteration using mean-squared error regression
to the teacher action. Optional point-level point-cloud augmentation is
disabled in the reported distillation runs.

\begin{table}[h]
\centering
\small
\caption{DAgger distillation and student-observation configuration.}
\label{tab:distill_hp}
\begin{tabular}{ll}
\toprule
\textbf{Parameter} & \textbf{Value} \\
\midrule
DAgger iterations $K$ & 30 \\
Rollout steps per iteration & 4096 \\
Initial mixing $\beta_0$ & 1.0 \\
Mixing decay & $0.85$ per iteration \\
Training epochs per iteration & 8 \\
Max buffer size & $200{,}000$ transitions \\
Reduced student observation & 417 \\
Scene points & 1024 \\
Hand keypoints & 6 \\
Tactile surface points & 25 \\
Point-cloud input shape & $1055\times5$ \\
PointNet output dimension & 64 \\
Student input dimension & 481 \\
Action dimension & 29 \\
\bottomrule
\end{tabular}
\end{table}

\section{Tactile Observation and Noise Model}
\label{app:tactile}

The state expert uses one scalar contact-force magnitude for each of the five
fingertips. The contact sensor maintains a short history, and the two most
recent force samples are averaged:
\begin{equation}
    f_t^{\mathrm{tac}}
=
\frac{1}{2}\left(
f_{t,0}^{\mathrm{raw}} + f_{t,1}^{\mathrm{raw}}
\right),
\end{equation}

The actor observation reserves 20 tactile-related dimensions: five force
magnitudes and 15 contact-position coordinates
($5$ fingertips $\times\,3$ coordinates). Contact-position observations are
disabled in the default configuration, and these 15 channels are explicitly
set to zero. Therefore, the effective tactile feedback used by the state
expert consists of five fingertip-force magnitudes.

During training, multiplicative Gaussian noise is applied to the force
magnitudes and the result is clipped to remain non-negative. Per-finger
dropout removes an entire fingertip signal, while a stochastic hold-last
operation models occasional sensing latency. Contact-position noise is only
relevant when contact-position observations are enabled.

\begin{table}[h]
\centering
\small
\caption{Tactile randomization parameters used during state-expert training.}
\label{tab:tactile_noise}
\begin{tabular}{ll}
\toprule
\textbf{Perturbation} & \textbf{Parameter} \\
\midrule
Force noise
    & $f\leftarrow\max(0,f(1+0.2\epsilon))$,
      $\epsilon\sim\mathcal{N}(0,1)$ \\
Per-finger dropout
    & $5\%$ \\
Hold-last probability
    & $0.005$ per element \\
Contact-position noise
    & $\sigma=3$\,mm (disabled by default) \\
Binary-contact mode
    & Disabled \\
\bottomrule
\end{tabular}
\end{table}

\section{Reward Details}
\label{app:reward}

The state-expert reward combines wrist tracking, absolute and wrist-relative
hand tracking, object tracking, contact shaping, action regularization, and
terminal success rewards. The implemented reward is organized as
\begin{equation}
\begin{aligned}
    r_t ={}&
    r^{\mathrm{wrist}}_t
    +2\,r^{\mathrm{hand,abs}}_t
    +r^{\mathrm{hand,rel}}_t
    +r^{\mathrm{object}}_t \\
    &+
    r^{\mathrm{contact}}_t
    +r^{\mathrm{action}}_t
    +r^{\mathrm{success}}_t
    +r^{\mathrm{collision}}_t ,
\end{aligned}
\end{equation}
where each group contains the weighted sub-terms summarized in
Table~\ref{tab:reward_w}. Tracking rewards are implemented as bounded
exponential functions of the corresponding position, orientation, or
velocity errors.

\begin{table}[h]
\centering
\small
\caption{Reward terms used for state-expert training. Absolute-hand tracking
terms are additionally multiplied by an outer group weight of $2.0$;
wrist and wrist-relative hand groups use outer weights of $1.0$.}
\label{tab:reward_w}
\begin{tabular}{llc}
\toprule
\textbf{Group} & \textbf{Component} & \textbf{Weight} \\
\midrule
Wrist
    & Position / rotation & $4.0 / 2.0$ \\
    & Linear / angular velocity & $0.1 / 0.05$ \\
\midrule
Absolute hand
    & Thumb / index / middle tip & $0.9 / 0.8 / 0.75$ \\
    & Pinky / ring tip & $0.6 / 0.6$ \\
    & Level-1 / level-2 tracking & $0.7 / 0.5$ \\
\midrule
Relative hand
    & Thumb / index / middle tip & $0.9 / 0.8 / 0.75$ \\
    & Pinky / ring tip & $0.6 / 0.6$ \\
    & Level-1 / level-2 tracking & $0.7 / 0.5$ \\
\midrule
Object
    & Position / rotation & $8.0 / 6.0$ \\
    & Linear / angular velocity & $0.1 / 0.4$ \\
    & Joint-velocity tracking & $0.1$ \\
\midrule
Contact
    & Fingertip force & $3.0$ \\
    & Approach shaping & $2.0$ \\
    & No-slip term & $1.5$ \\
\midrule
Action
    & Action-rate term & $+0.1$ \\
    & Arm action-rate term & $-0.15$ \\
\midrule
Terminal
    & Final position / rotation / approach & $30.0 / 5.0 / 1.0$ \\
\bottomrule
\end{tabular}
\end{table}

The approach shaping term uses the minimum fingertip-to-object distance
$d^{\min}_t$:
\begin{equation}
    r^{\mathrm{approach}}_t
    =
    \frac{1}{1+5d^{\min}_t}.
\end{equation}
The environment additionally applies an arm-collision penalty when relevant.
Unlike the simplified reward abstraction, the implemented reward does not
contain a separate joint-limit penalty; joint limits are instead enforced by
the controller.

\section{Domain Randomization}
\label{app:dr}

Table~\ref{tab:dr} summarizes the main domain-randomization parameters used
during state-expert training. Tactile randomization is described separately
in Appendix~\ref{app:tactile}. Object-pose perturbations emulate errors in the
real-world pose-estimation pipeline through a combination of frame-wise
noise, episode-wise bias, latency, and dropout.

\begin{table}[h]
\centering
\small
\caption{Domain-randomization parameters used during state-expert training.}
\label{tab:dr}
\begin{tabular}{lll}
\toprule
\textbf{Parameter} & \textbf{Range} & \textbf{Purpose} \\
\midrule
Hand PD stiffness
    & $\times[0.5,2.0]$ & Hand dynamics \\
Hand PD damping
    & $\times[0.5,2.0]$ & Hand dynamics \\
Arm PD gains
    & $\times[0.8,1.2]$ & Arm dynamics \\
Object mass
    & $[0.01,0.15]$\,kg & Object dynamics \\
Object CoM offset
    & $\pm0.02$\,m per axis & Object dynamics \\
Friction
    & $\times[1.0,2.5]$ & Contact dynamics \\
Action delay
    & $0$--$3$ steps & Control latency \\
\midrule
Object-pose position noise
    & $\sigma=8$\,mm & Frame-wise error \\
Object-pose rotation noise
    & $\sigma=0.06$\,rad & Frame-wise error \\
Object-pose position bias
    & $\mathcal{U}(-0.05,0.05)$\,m per axis & Episode-wise bias \\
Object-pose rotation bias
    & $\sigma=0.05$\,rad & Episode-wise bias \\
Object-pose latency
    & $2$ steps & Perception latency \\
Object-pose dropout
    & $2\%$ per step & Hold-last observation \\
Tactile
    & See App.~\ref{app:tactile} & Sensor mismatch \\
\bottomrule
\end{tabular}
\end{table}

\section{Observation Space}
\label{app:obs}

The state expert uses a 557-dimensional actor observation and an additional
148-dimensional privileged critic state. The complete critic input therefore
contains $557+148=705$ dimensions.

\begin{table}[h]
\centering
\small
\caption{State-expert actor observation layout ($557$ dimensions).}
\label{tab:obs}
\begin{tabular}{llr}
\toprule
\textbf{Group} & \textbf{Component} & \textbf{Dim.} \\
\midrule
Proprioception
    & Hand joint positions & 22 \\
    & $\cos(\mathbf{q}^{\mathrm{hand}})$ & 22 \\
    & $\sin(\mathbf{q}^{\mathrm{hand}})$ & 22 \\
    & Reserved wrist-position channel & 3 \\
    & Wrist quaternion & 4 \\
    & Wrist linear velocity & 3 \\
    & Wrist angular velocity & 3 \\
    & \textit{Subtotal} & \textit{79} \\
\midrule
Wrist reference
    & Target-current wrist position & 3 \\
    & Target wrist linear velocity & 3 \\
    & Target-current linear velocity & 3 \\
    & Target wrist quaternion & 4 \\
    & Target-current wrist quaternion & 4 \\
    & Target angular velocity & 3 \\
    & Target-current angular velocity & 3 \\
    & \textit{Subtotal} & \textit{23} \\
\midrule
Hand reference
    & Target-current keypoint positions & 96 \\
    & Target keypoint velocities & 96 \\
    & Target-current keypoint velocities & 96 \\
    & \textit{Subtotal} & \textit{288} \\
\midrule
Task reference
    & Target object position & 3 \\
    & Target object quaternion & 4 \\
    & Fingertip-to-object distances & 5 \\
\midrule
Object geometry
    & BPS object encoding & 128 \\
\midrule
Tactile
    & Fingertip force magnitudes & 5 \\
    & Reserved contact-position channels & 15 \\
\midrule
Current object pose
    & Noisy position + quaternion & 7 \\
\midrule
\multicolumn{2}{l}{\textbf{Total}} & \textbf{557} \\
\bottomrule
\end{tabular}
\end{table}

The three-dimensional absolute wrist-position slot and the 15-dimensional
contact-position block are retained for a fixed observation layout but are
zeroed in the reported configuration. The retargeted wrist and hand motion
reference occupies $23+288=311$ dimensions.

\begin{table}[h]
\centering
\small
\caption{Privileged critic-state layout ($148$ dimensions).}
\label{tab:critic_obs}
\begin{tabular}{llr}
\toprule
\textbf{Group} & \textbf{Component} & \textbf{Dim.} \\
\midrule
Current state
    & Hand joint velocities & 22 \\
    & Physical-parameter slots & 5 \\
    & Ground-truth object position & 3 \\
    & Ground-truth object quaternion & 4 \\
    & Ground-truth object linear/angular velocity & 6 \\
    & \textit{Subtotal} & \textit{40} \\
\midrule
Five-step future
    & Target object positions & 15 \\
    & Target object quaternions & 20 \\
    & Target linear velocities & 15 \\
    & Target angular velocities & 15 \\
    & Fingertip-to-object distances & 25 \\
    & \textit{Subtotal} & \textit{90} \\
\midrule
Current relation
    & Target-current object-state delta & 13 \\
    & Current fingertip-object distances & 5 \\
    & \textit{Subtotal} & \textit{18} \\
\midrule
\multicolumn{2}{l}{\textbf{Total}} & \textbf{148} \\
\bottomrule
\end{tabular}
\end{table}

The five physical-parameter slots are retained by the fixed privileged-state
layout. The semantic critic description above therefore focuses on the
privileged quantities actively used for state and future-trajectory
conditioning.

\paragraph{Student observation.}
The student removes three object-information blocks from the teacher actor
observation: the 128-dimensional BPS encoding, the 5-dimensional reference
fingertip-to-object distances, and the 7-dimensional noisy current-object
pose estimate. This yields a 417-dimensional vector observation:
\begin{equation}
    557 - 128 - 5 - 7 = 417.
\end{equation}
The demonstration target object pose remains part of the retained motion
reference.

The student additionally receives a visual-tactile point cloud composed of
1024 scene points, six hand keypoints, and 25 fingertip-surface tactile
points. Each point contains three spatial coordinates, one scalar type
indicator, and one scalar force channel, yielding
\begin{equation}
    (1024+6+25)\times(3+1+1)
    =
    1055\times5.
\end{equation}
A shared PointNet maps this cloud to a 64-dimensional feature, producing the
final student input dimension
\begin{equation}
    417+64=481.
\end{equation}
The vector observation retains the five scalar fingertip-force signals,
while the point-cloud branch additionally spatializes these forces over the
25 tactile surface points.

\section{Sim-to-Real Alignment Details}
\label{app:sim2real}

\paragraph{Arm-dynamics calibration.}
We replay sinusoidal trajectories ($0.2$\,Hz, amplitude $0.3$\,rad) on each arm
joint independently, recording commanded and achieved positions on both the
real FR3 and in simulation. We adjust simulated damping coefficients for
joints $1$--$3$ (reduced by approximately $40\%$) until the cross-correlation
exceeds $0.996$ and the estimated lag is below $10$\,ms.

\paragraph{Deploy safety.}
The deployment controller implements joint-position limits with margin,
joint-velocity limits, arm- and hand-action delta limits, and a two-phase
reset ramp (current pose $\to$ home pose $\to$ demo start) to avoid sudden
motions at episode boundaries.

\paragraph{Control-loop timing.}
The $30$\,Hz budget is approximately $33$\,ms per step: depth capture
($\sim5$\,ms), forward kinematics ($\sim2$\,ms), policy inference
($\sim3$\,ms), ROS\,2 publish ($\sim1$\,ms), and Sharpa SDK communication
($\sim1$\,ms via a background thread), with margin for scheduling jitter.

\end{document}